\RequirePackage{amsmath}
\documentclass[runningheads,a4paper]{llncs}\usepackage[]{graphicx}\usepackage[]{color}
\makeatletter
\def\maxwidth{ %
  \ifdim\Gin@nat@width>\linewidth
    \linewidth
  \else
    \Gin@nat@width
  \fi
}
\makeatother

\definecolor{fgcolor}{rgb}{0.345, 0.345, 0.345}

\usepackage{framed}
\makeatletter
 {\par\unskip\endMakeFramed%
 \at@end@of@kframe}
\makeatother

\definecolor{shadecolor}{rgb}{.97, .97, .97}
\definecolor{messagecolor}{rgb}{0, 0, 0}
\definecolor{warningcolor}{rgb}{1, 0, 1}
\definecolor{errorcolor}{rgb}{1, 0, 0}
\newenvironment{knitrout}{}{} % an empty environment to be redefined in TeX

\usepackage{alltt}

\usepackage[%
rm={oldstyle=false,proportional=true},%
sf={oldstyle=false,proportional=true},%
tt={oldstyle=false,proportional=true,variable=true},%
qt=false%
]{cfr-lm}
\usepackage{graphicx}

\usepackage{paralist}
\usepackage[caption=false]{subfig}
\usepackage[T1]{fontenc}

\usepackage[math]{blindtext}

\usepackage{csquotes}

\usepackage{microtype}

\usepackage{url}
\makeatletter
\def\Url@twoslashes{\mathchar`\/\@ifnextchar/{\kern-.2em}{}}
\g@addto@macro\UrlSpecials{\do\/{\Url@twoslashes}}
\makeatother
\makeatletter
\g@addto@macro{\UrlBreaks}{\UrlOrds}
\makeatother

\usepackage[
bookmarks=false,
breaklinks=true,
colorlinks=true,
linkcolor=black,
citecolor=black,
urlcolor=black,
pdfpagelayout=SinglePage,
pdfstartview=Fit
]{hyperref}
\usepackage[all]{hypcap}

\usepackage[capitalise]{cleveref}
\crefname{section}{Sect.}{Sect.}
\Crefname{section}{Section}{Sections}
\crefname{figure}{Fig.}{Fig.}
\Crefname{figure}{Figure}{Figures}

\usepackage{xspace}
\DeclareFontFamily{U}{MnSymbolC}{}
\DeclareSymbolFont{MnSyC}{U}{MnSymbolC}{m}{n}
\DeclareFontShape{U}{MnSymbolC}{m}{n}{
    <-6>  MnSymbolC5
   <6-7>  MnSymbolC6
   <7-8>  MnSymbolC7
   <8-9>  MnSymbolC8
   <9-10> MnSymbolC9
  <10-12> MnSymbolC10
  <12->   MnSymbolC12%
}{}
\usepackage{amsmath}
\DeclareMathSymbol{\powerset}{\mathord}{MnSyC}{180}

\DeclareMathOperator*{\argmax}{arg\,max}
\IfFileExists{upquote.sty}{\usepackage{upquote}}{}
\begin{document}

\title{The Peaking Phenomenon\\ in Semi-supervised Learning}

\author{Jesse H. Krijthe\inst{1,2} \and Marco Loog\inst{1,3}}

\institute{
Pattern Recognition Laboratory, Delft University of Technology\\
\and
Department of Molecular Epidemiology, Leiden University Medical Center\\
\and
The Image Section, University of Copenhagen\\
\email{jkrijthe@gmail.com}
}
			
\maketitle

\begin{abstract}
For the supervised least squares classifier, when the number of training objects is smaller than the dimensionality of the data, adding more data to the training set may first increase the error rate before decreasing it.  This, possibly counterintuitive, phenomenon is known as peaking. In this work, we observe that a similar but more pronounced version of this phenomenon also occurs in the semi-supervised setting, where instead of labeled objects, unlabeled objects are added to the training set. We explain why the learning curve has a more steep incline and a more gradual decline in this setting through simulation studies and by applying an approximation of the learning curve based on the work by Raudys \& Duin.
\end{abstract}


\section{Introduction}
In general, for most classifiers, classification performance is expected to improve as more labeled training examples become available. The dipping phenomenon is one exception to this rule, showing for specific combinations of datasets and classifiers that error rates can actually increase with increasing numbers of labeled data \cite{Loog2012}.  For the least squares classifier and some other classifiers, the \emph{peaking phenomenon} is another known exception. In this setting, the classification error may first increase, after which the error rate starts to decrease again as we add more labeled training examples. The term peaking comes from the form of the learning curve: an example of which is displayed in \cref{fig:peaking}. 

\begin{knitrout}
\definecolor{shadecolor}{rgb}{0.969, 0.969, 0.969}\color{fgcolor}\begin{figure}

{\centering \includegraphics[width=\maxwidth]{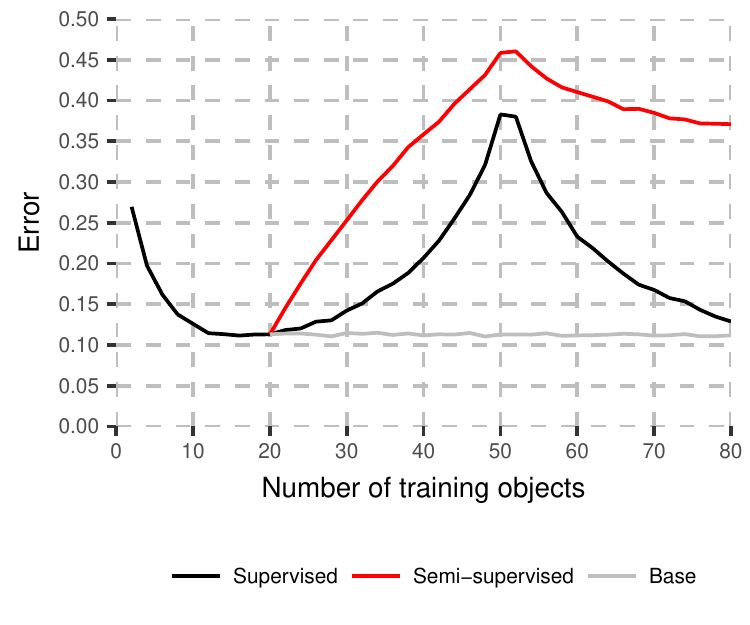} 

}

\caption{Empirical learning curves for the supervised least squares classifier (Eq.~\eqref{eq:solutionsupervised}) where labeled data is added and the semi-supervised least squares classifier (Eq.~\eqref{eq:solutionsemisupervised}) which uses $10$ labeled objects per class and the remaining objects as unlabeled objects. ``Base'' corresponds to the performance of the classifier that uses the first $10$ labeled objects for each class, without using any additional objects. Data are generated from two Gaussians in $50$ dimensions, with identity covariance matrices and a distance of $4$ between the class means.}\label{fig:peaking}
\end{figure}

\end{knitrout}

The term `peaking' is inspired by a different peaking phenomenon described by \cite{Hughes1968} (see also \cite{Jain1982}), who studies the phenomenon that the performance of many classifiers peaks for a certain number of features and then decreases as more features are added. In this work we consider a different peaking phenomenon that occurs when the number of training objects is increased, and the peak does not refer to a peak in performance, but a peak in terms of the classification error, after which performance starts increasing again. While this type of peaking also shows up in feature curves, where we increase the number of features, we focus on learning curves in terms of the number of training objects because it relates more closely to the question whether unlabeled data should be used at all in the semi-supervised setting.

The peaking phenomenon considered here is observed in \cite{Duin1995,Skurichina1996,Opper1995,Duin2000,Opper2001} for various classifiers in the supervised learning setting and \cite{Duin1995,Skurichina1999,Duin2000,Opper2001} additionally describe different ways to get rid of this unwanted behaviour, notably, by only considering a subset of relevant objects, by adding regularization to the parameter estimation, adding noise to objects, doing feature selection, or by injecting random features.

While this peaking phenomenon has been observed for the least squares classifier when the amount of labeled data is increased, we find similar but worse behaviour in the semi-supervised setting.  Following the work in \cite{Duin1995} and \cite{Fan2008}, we study a particular semi-supervised adaptation of the least squares classifier in greater depth. An example of the actual behaviour is shown in \cref{fig:peaking}. When the amount of labeled objects remains fixed ($20$ in the figure) while we increase the amount of unlabeled data used by this semi-supervised learner, the peaking phenomenon changes in two ways: the error increases more rapidly when unlabeled data is added than when labeled data is added and after the peak the error decreases more slowly than when labeled data is added. The goal of this work is to describe and explain these effects. More specifically, we attempt to answer two questions:
\begin{enumerate}
\item What causes the performance in the semi-supervised setting to deteriorate faster than in the supervised case?
\item If we increase the amount of unlabeled data, will the performance of the semi-supervised learner converge to an error rate below the error rate of the supervised learner that does not take the additional unlabeled data into account?
\end{enumerate}
To answer these questions, we first revisit the supervised peaking phenomenon and explain its causes in \cref{section:supervisedpeaking}. In \cref{section:semi-supervised} we show how the results from \cref{section:supervisedpeaking} relate to the least squares classifier and how we specifically adapt this classifier to the semi-supervised setting. In \cref{section:incline,section:decline} we attempt to answer our two questions in two ways: firstly by adapting the learning curve approximation of Raudys \& Duin \cite{Raudys1998} and secondly through simulation studies. We end with an investigation of the semi-supervised peaking phenomenon on some benchmark datasets.

\section{Supervised Peaking} \label{section:supervisedpeaking}
Raudys \& Duin \cite{Raudys1998} attempt to explain the peaking phenomenon in the supervised case by constructing an asymptotic approximation of the learning curve and decomposing this approximation into several terms that explain the effect of adding labeled data on the learning curve. The classifier they consider is the Fisher linear discriminant, whose normal to the decision boundary is defined as the direction that maximizes the between-class variance while minimizing the within-class variance:
\begin{equation} \label{eq:fisherobjective}
\argmax_{\vec{w}} \frac{(\vec{w}^\top \vec{m}_1 - \vec{w}^\top \vec{m}_2)^2}{\vec{w}^\top W \vec{w}} \, ,
\end{equation}
where $\vec{m}_c$ is the sample mean of class $c$ and $W=\tfrac{1}{n} \sum_{c=1}^2 \sum_{i=1}^{N_c} (\vec{x}_{ci} - \vec{m}_c)(\vec{x}_{ci} - \vec{m}_c)^\top$ is the sample within-class scatter matrix. The solution is given by
\begin{equation} \label{eq:Wformulation}
\vec{w} = W^{-1} (\vec{m}_1-\vec{m}_2) \, .
\end{equation}
The intercept (or threshold value) that we consider in actual classification is right in between the two class means:
$-\frac{1}{2}(\vec{m}_1+\vec{m}_2)^\top \vec{w}$.  The peaking phenomenon occurs when $n=2N<p$, where $N$ is the number of (labeled) objects per class and $p$ is the dimensionality of the data. In this case, a pseudo-inverse needs to be applied instead of the regular inverse of $W$. This is equivalent to removing directions with an eigenvalue of $0$ and training the classifier in a lower dimensional subspace, a subspace whose dimensionality increases as more training data is added.

The goal of the analysis in \cite{Raudys1998} is to construct an approximation of the learning curve, which decomposes the error into different parts. These parts relate the observed peaking behaviour to different individual effects of increasing the number of training objects. To do this they construct an asymptotic approximation where both the dimensionality and the number of objects grows to infinity. An important assumption in the derivation, and the setting we also consider in our analysis, is that the data are generated from two Gaussian distributions corresponding to two classes, with true variance $\mathbf{I}$ and a Euclidean distance between the true means $\vec{\mu}_1$ and $\vec{\mu}_2$ of $\delta$. Lastly, objects are sampled in equal amounts from both classes.

The approximation of the learning curve is then given by\footnote{While going through the derivation we found a different solution than the one reported in \cite{Raudys1998}, which renders the last term in the formulation independent of $N$. This slightly changes the expressions in the explanation of the peaking behaviour.}
% $$
% R(N,p,\delta_d,\gamma,\mu_4) = \Phi \left\{ - \frac{\delta}{2} \sqrt{(1+\gamma^2)(1+ \frac{1}{N} (1+\frac{2 p}{\delta_d^2})+\frac{p}{\delta^2_d N^2})+\gamma^2 \frac{\mu_4}{4 \delta_d^2}}^{-1}\right\} \,.
% $$
% When we substitute estimates of $\delta_d^2$ and $\mu_4$ into this formulation
$$e(N,p,\delta) = \Phi \left\{ - \frac{\delta}{2} T_r \sqrt{(1+\gamma^2)T_\mu+\gamma^2 \frac{3 \delta^2}{4 p}}^{-1}\right\} \,,$$
where $\Phi$ is the cumulative distribution function of a standard normal distribution and $N$ is the number of objects per class.  The main quantities introduced are $T_\mu$, $T_r$, and $\gamma$ and \cite{Raudys1998} notes that the approximation of the learning curve can be broken down to depend on exactly these three quantities all with their own specific interpretation:
%$$T_\mu=1+\frac{1}{N} + \frac{4 p^2}{\delta^2 N^2} + \frac{2 p^2}{\delta^3 N^3}$$
%which relates to how well we can estimate the means of the classes, $T_r=\sqrt{\frac{N}{2 p}}$ which relates to the reduction in features brought about by using  the pseudo-inverse and two terms relating to the estimation of the eigenvalues, $\gamma=\frac{\sqrt{V_d}}{E_d}$ and $T_{\textrm{eig}}=\frac{3 \delta^2 N}{8 p^2}$.
$$T_\mu=1+ \frac{1}{N} + \frac{2 p^2}{\delta^2 (2N-2) N}+\frac{p^2}{\delta^2 (2N-2) N^2}\,,$$
relates to how well we can estimate the means, $T_r=\sqrt{\frac{2 N-2}{p}}$ relates to the reduction in features brought about by using the pseudo-inverse and $\gamma$ is a term related to the estimation of the eigenvalues or $W$.  The $T_\mu$ and $T_r$ terms lead to a decrease in the error rate as $N$, the number of objects per class increases. This is caused by the improved estimates of the means and the increasing dimensionality. The $\gamma$ term increases the generalization error as $N$ increases, which is caused by the fact that the smallest eigenvalues are difficult to accurately estimate but can have a large effect on the computation of the pseudo-inverse.

When $n>p$ the pseudo-inverse is no longer necessary and other approximations of the learning curve can be applied. The comparison of these approximations in \cite{Wyman1990} shows that the approximation
$$
e(N,p,\delta) = \Phi \left\{ - \frac{\delta}{2}  \sqrt{T_\mu T_\Sigma}^{-1}\right\} \,,
$$
with $T_\mu=1+\frac{2 p}{\delta^2 N}$ and $T_\Sigma=1+\frac{p}{2 N - p}$ works reasonably well. The former term again relates to the estimation of the means while the latter term relates to the estimation of the within scatter matrix $W$.  \cref{fig:peaking-asymptotic1} shows these approximations and the empirical learning curve on a simple dataset with 2 Gaussian classes, with a distance between the means of $\delta=4.65$.

\begin{knitrout}
\definecolor{shadecolor}{rgb}{0.969, 0.969, 0.969}\color{fgcolor}\begin{figure}
\subfloat[Within Scatter\label{fig:peaking-asymptotic1}]{\includegraphics[width=.49\linewidth]{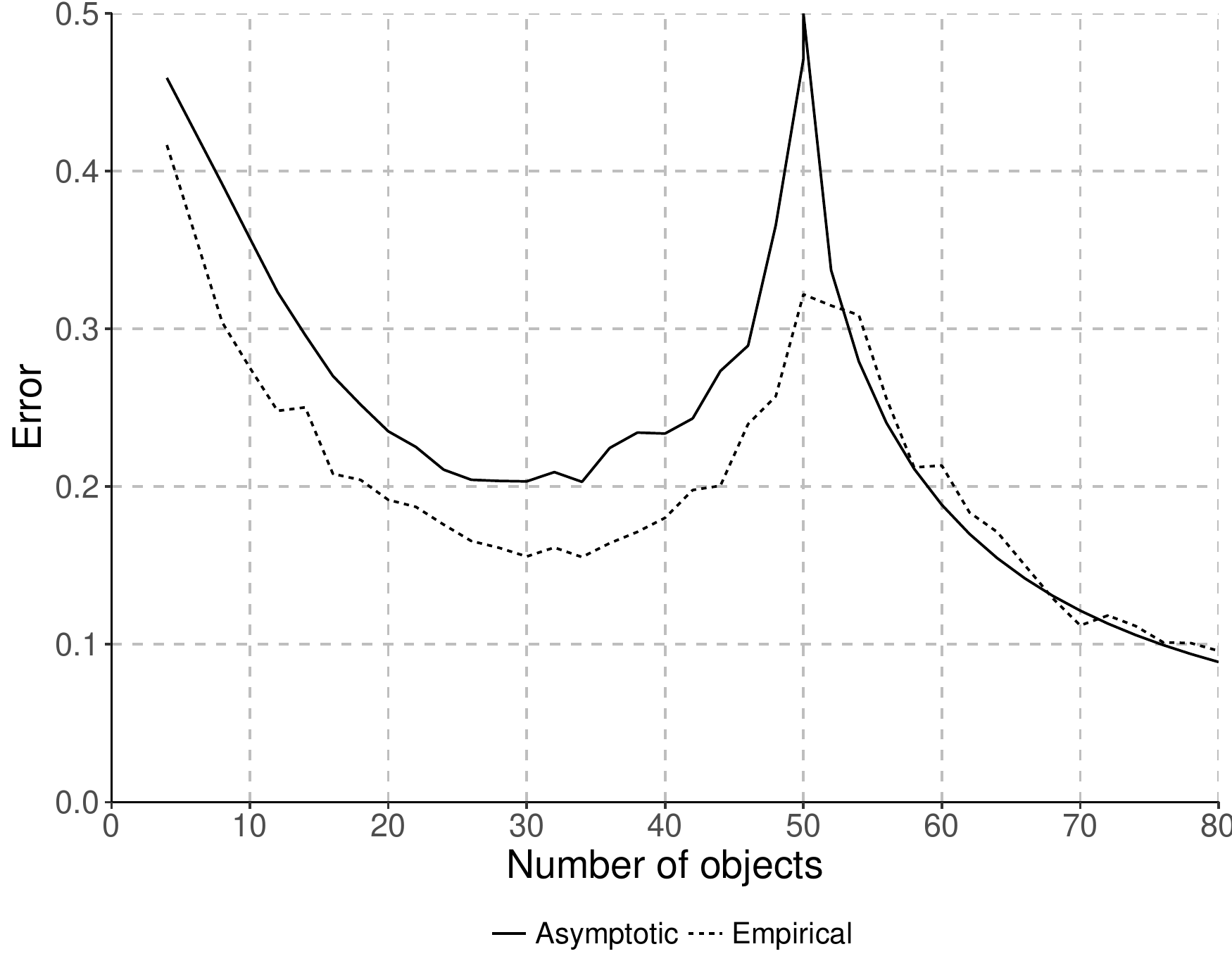} }
\subfloat[Total Scatter\label{fig:peaking-asymptotic2}]{\includegraphics[width=.49\linewidth]{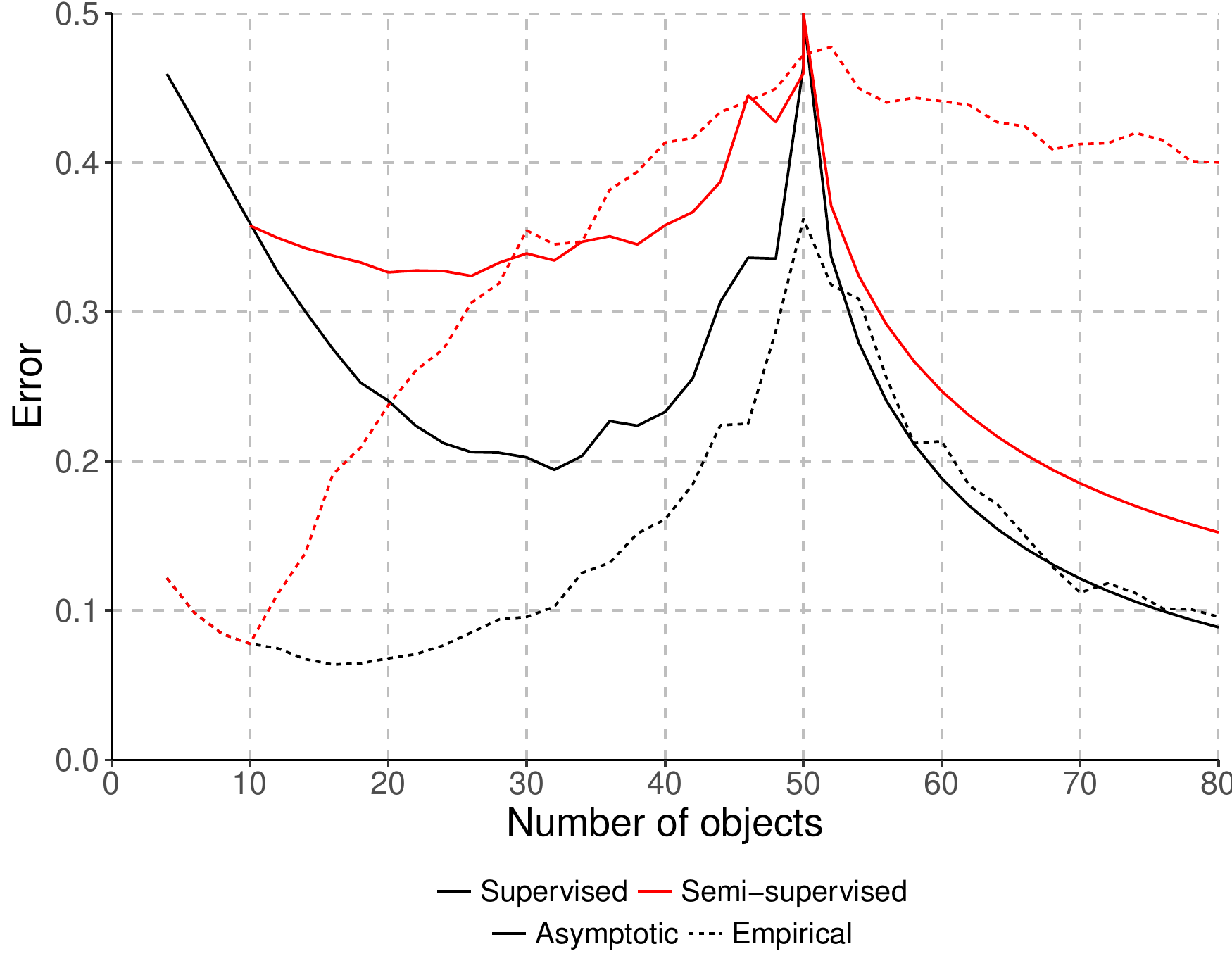} }\caption{Empirical learning curves and their asymptotic approximations for different classifiers. (a) Supervised learning curve corresponding to the formulation in Eq.~\eqref{eq:Wformulation}. (b) Supervised and semi-supervised learning curves corresponding to the formulations in Eqs.~\eqref{eq:Tformulation}, \eqref{eq:solutionsupervised} and \eqref{eq:solutionsemisupervised}. Semi-supervised uses $5$ labeled objects per class and the rest as unlabeled objects.}\label{fig:peaking-asymptotic}
\end{figure}

\end{knitrout}

\section{Semi-supervised Classifier} \label{section:semi-supervised}
Unfortunately for our analysis, the classifier studied by Raudys \& Duin does not correspond directly to the least squares classifier we wish to study, nor is it directly clear how their classifier can be extended to the semi-supervised setting.
% \cite{Fan2008} propose a slightly different formulation of the objective function that does allow for the incorporation of unlabeled data and which corresponds to the least squares classifier. It is based on the observation that the total scatter matrix, $T=\tfrac{1}{n}\sum_{i=1}^n (\vec{x}_i - \vec{m})(\vec{x}_i - \vec{m})^\top$, with $\vec{m}$ the overall mean of the data, can be decomposed as $T=B+W$. $B=\tfrac{1}{2}\sum_{c=1}^2 (\vec{m}_c - \vec{m})(\vec{m}_c - \vec{m})^\top$ is the between class scatter matrix. Because of this equality maximizing
% $$\frac{\vec{w}^\top T \vec{w} }{\vec{w}^\top W \vec{w} }$$
% has the same maximum as the objective in Equation \eqref{eq:fisherobjective} and we can write the solution as:
We therefore consider a slightly different version in which we follow \cite{Duin1995} and \cite{Fan2008}:
\begin{equation}
\vec{w} = T^{-1} (\vec{m}_1-\vec{m}_2) \label{eq:Tformulation} \, .
\end{equation}
This leads to the same classifier as Equation \eqref{eq:Wformulation} when $n>p$ \cite{Duin1995}. Moreover, when the data are centered ($\vec{m}=\vec{0}$) and the class priors are exactly equal it is equivalent to the solution of the least squares classifier, which minimizes the squared loss $(\vec{x}_i^\top \vec{w} - y_i)^2$ and whose solution is given by
\begin{equation}
\vec{w} = (\mathbf{X}^\top \mathbf{X})^{-1} \mathbf{X}^\top \vec{y} \, , \label{eq:solutionsupervised}
\end{equation}
where $\vec{y}$ is a vector containing a numerical encoding of the labels and $\mathbf{X}$ is the $L \times p$ design matrix containing the $L$ labeled feature vectors $\vec{x}_i$.

While Eq.~\eqref{eq:Tformulation} is equivalent to Eq.~\eqref{eq:Wformulation} when $n>p$, this solution is not necessarily the same in the scenario where $n<p$ (compare the dashed black lines in \cref{fig:peaking-asymptotic1} and \cref{fig:peaking-asymptotic2}). This makes it impossible to apply the results from \cite{Raudys1998} directly to get a quantitatively good estimator for the learning curve. Moreover, their proof is not easily adapted to this new classifier. This is caused by dependencies that are introduced between the total scatter matrix $T$ (which is proportional to $\mathbf{X}^\top \mathbf{X}$ in case $\vec{m}=\vec{0}$) and the mean vectors $\vec{m}_c$ that complicate the derivation of the approximation. Their result does, however, offer a qualitative explanation of the peaking phenomenon in the semi-supervised setting---as we will see in \cref{section:inclineasymptotic}.

How then do we adapt the least squares classifier to the semi-supervised setting? Reference \cite{Fan2008} proposes to update $T$, which does not depend on the class labels, based on the additional unlabeled data. Equivalently, in the least squares setting, \cite{Shaffer1991} studies the improvement in the least squares classifier by plugging in a better estimator of the covariance term, $\mathbf{X}^\top \mathbf{X}$, which is equivalent to the update proposed by \cite{Fan2008}. We define our semi-supervised least squares classifier as this update:
\begin{equation}
\vec{w} = (\tfrac{L}{L+U} \mathbf{X}_\textrm{e}^\top \mathbf{X}_\textrm{e})^{-1} \mathbf{X}^\top \vec{y} \,. \label{eq:solutionsemisupervised}
\end{equation}
This is the semi-supervised learner depicted in \cref{fig:peaking}. Here $L$ is the number of labeled objects, $U$, the number of unlabeled objects and $\mathbf{X}_\textrm{e}$ the $(L+U) \times p$ design matrix containing all the feature vectors. The weighting $\tfrac{L}{L+U}$ is necessary because $\mathbf{X}_\textrm{e}^\top \mathbf{X}_\textrm{e}$ is essentially a sum over more objects than $\mathbf{X}^\top \vec{y}$, which we have to correct for.

\section{Why peaking is more extreme under semi-supervision} \label{section:incline}
One apparent feature of the semi-supervised peaking phenomenon is that before the peak occurs, the learning curve rises more steeply when unlabeled data are added vs. when labeled data are.

\subsection{Asymptotic Approximation} \label{section:inclineasymptotic}
To explain this behaviour using the learning curve approximation, we hold the term that relates to the increased accuracy of the estimate of the means, $T_\mu$, constant and consider the change in the approximation. As we noted before, the learning curve approximation is for a slightly different classifier, yet it might offer a qualitative insight as to the effect of only adding unlabeled data. Looking at the resulting curve in \cref{fig:peaking-asymptotic2}, we indeed see that the semi-supervised approximation rises more quickly than the supervised approximation due to the lack of labeled data to improve the estimates of the mean. After the peak we see that the curve drops off less quickly for the same reason. The approximation, however, is not a very accurate reflection of the empirical learning curve.

\subsection{Simulation of Contributions}
\begin{knitrout}
\definecolor{shadecolor}{rgb}{0.969, 0.969, 0.969}\color{fgcolor}\begin{figure}
\includegraphics[width=\maxwidth]{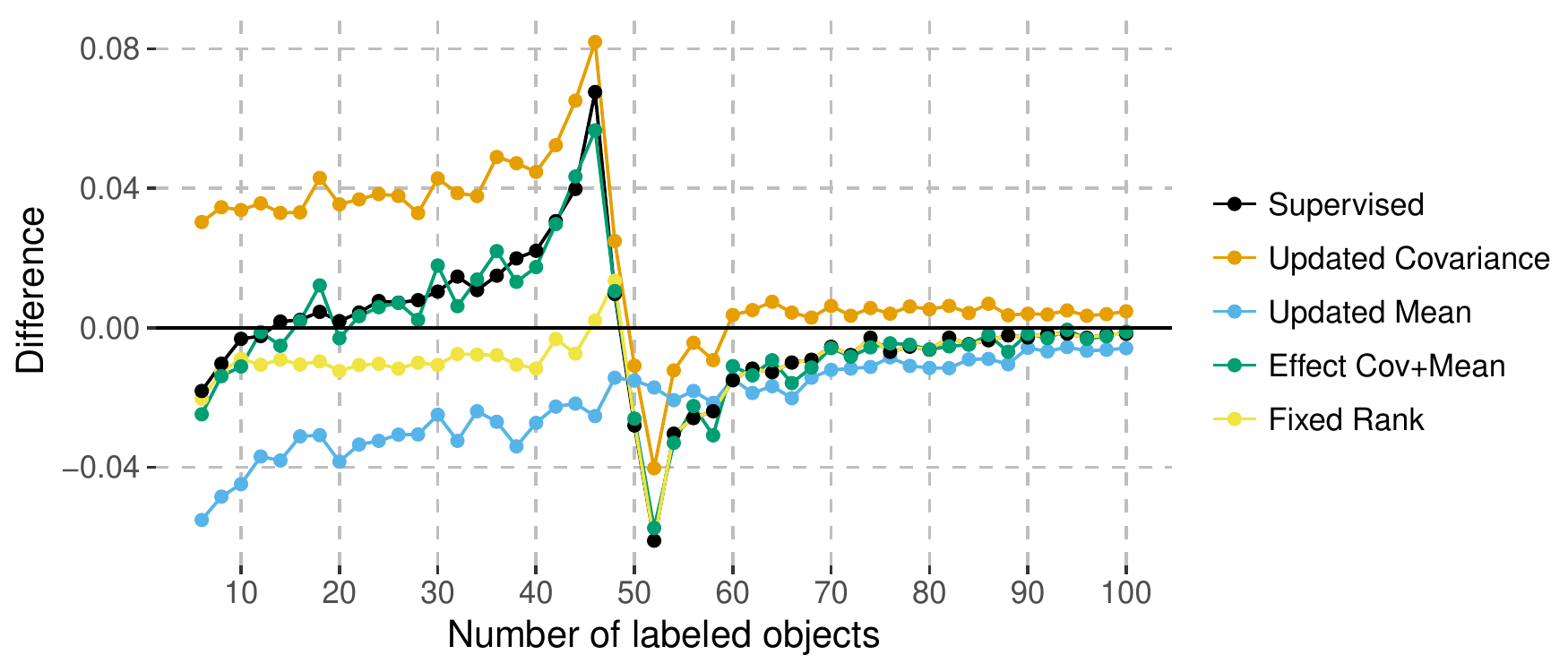} \caption[Average gain in error rate by adding $2$ additional objects to the training set, either in a supervised way, by adding labeled objects, or by only using them to improve the estimator of the total scatter (covariance)]{Average gain in error rate by adding $2$ additional objects to the training set, either in a supervised way, by adding labeled objects, or by only using them to improve the estimator of the total scatter (covariance).}\label{fig:contributions}
\end{figure}

\end{knitrout}
Because the approximation used does not approximate the empirical learning curve very well, the question remains whether the lack of the updating of the means based on new data fully explains the increase in the semi-supervised learning curve over the supervised learning curve. To explore this, we decompose the change in the supervised learning curve into separate components by calculating the change in the error rate from adding data to improve respectively the estimator of the total covariance, $T$, the means or both at the same time. The result is shown in \cref{fig:contributions}.

To do this we compare the difference in error of the semi-supervised classifier that has two additional unlabeled objects available to the supervised classifier that does not have these unlabeled data available. We see that adding these objects typically increases the error rate when $n<p$.  We then compare the error of the supervised classifier to the one where we remove $2$ labels and the classifier where we do not remove these labels. By negating this difference we get the value of having two additional labels. We see that for this dataset this effect always decreases the error.  Adding up the effect of adding unlabeled objects to the effect of having additional labels, we find this approximates the total effect of adding labeled objects very well. It seems, therefore, that in the semi-supervised setting, by not having additional labels, the positive effect of these labels as shown \cref{fig:contributions} is removed, explaining the difference between the supervised and semi-supervised setting.

It is also clear from these results that peaking is caused by the estimation of the inverse of the covariance matrix, which leads to an increase in the error before $n>p$. To understand why this happens, consider the ``Fixed rank'' curve in \cref{fig:contributions}. This curve shows the change in terms of the error rate when we add two labeled objects but leave the rank of the covariance matrix unchanged during the calculation of the inverse, merely considering the largest $n$ eigenvectors of the newly obtained covariance matrix that was estimated using $n+2$ objects. Since this tends to decrease the error rate, the error increase for the other curve may indeed stem from the actual growth of the rank. Especially when $n$ is close to $p$, the eigenvalues of the dimensions that are added by increasing the rank become increasingly hard to estimate. This is similar to the $\gamma$ term in the approximation, which captures the difficulty of estimating the eigenvalues for these directions.

\begin{knitrout}
\definecolor{shadecolor}{rgb}{0.969, 0.969, 0.969}\color{fgcolor}\begin{figure}
\includegraphics[width=\maxwidth]{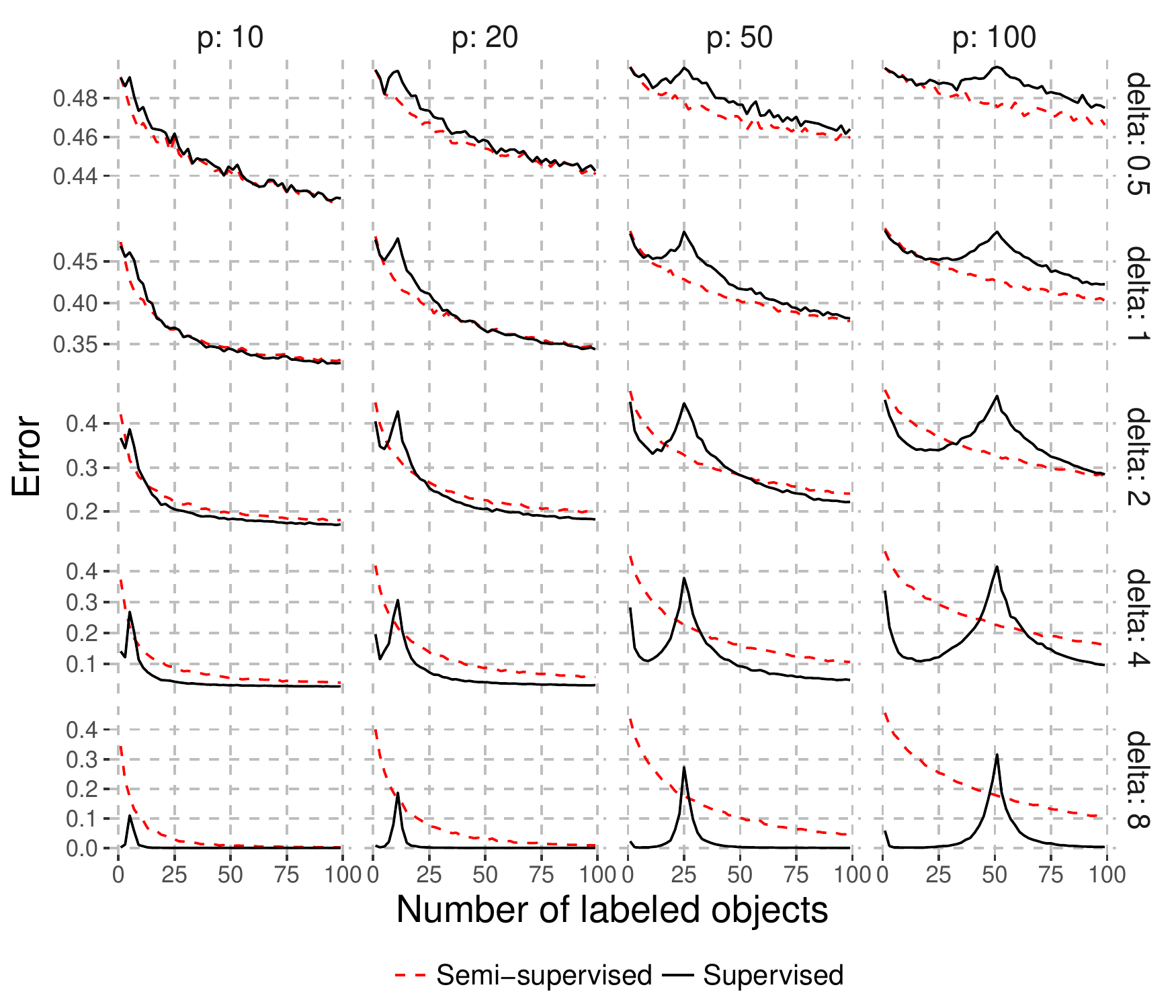} \caption[Learning curves for the supervised learner and the semi-supervised learner with infinite amounts of unlabeled data for different dimensionalities, $p$, and distances between the means, $\delta$]{Learning curves for the supervised learner and the semi-supervised learner with infinite amounts of unlabeled data for different dimensionalities, $p$, and distances between the means, $\delta$.}\label{fig:infinitedata}
\end{figure}

\end{knitrout}

\section{Convergence to a better solution than the base learner?} \label{section:decline}

The slow decline of the error rate after the peak in the learning curve begs the question whether the semi-supervised learner's error will ever drop below the error of the original supervised learner. If not, it would be worthwhile to refrain from using the semi-supervised learner in these settings.  The approximation in \cref{fig:peaking-asymptotic2} indicates that the learning curve will decline more slowly when $n>p$ when unlabeled data are added. From this approximation, however, it is not clear if and under which circumstances the error of the semi-supervised classifier will improve over the base learner if larger amounts of unlabeled data become available.

To investigate this issue we consider, for the two-class Gaussian problem with different dimensionalities, $p$, and different distances between the means, $\delta$, whether adding infinite unlabeled data improves over the supervised learner, for different amounts of limited labeled data. We can simulate this by setting the true means as $\vec{\mu}_1=-\tfrac{\delta}{2 \sqrt{p}}\vec{1}$ and $\vec{\mu}_2=+\tfrac{\delta}{2 \sqrt{p}}\vec{1}$. In this case, when the amount of unlabeled data increases, the total scatter matrix will converge to
$$
T = \vec{I} + \vec{1} \vec{1}^\top \frac{1}{4} \frac{\delta^2}{p} \,.
$$
Using this we can calculate the semi-supervised classifier based on an infinite unlabeled sample and with a finite amount of labeled data. The results are shown in \cref{fig:infinitedata}.

We observe that the dimensionality of the data does not have a large effect on whether the semi-supervised learner can outperform the supervised learner. It merely shifts the peak while qualitatively the differences between the supervised and semi-supervised curves remain the same. If we decrease the Bayes error by moving the means of the classes further apart, however, there are clear changes. For small distances between the means, the semi-supervised learner generally does increase performance for a larger range of sizes of the labeled set, while for larger distances this is no longer the case and the semi-supervised solution is typically worse than the supervised solution that does not take the unlabeled data into account.

\section{Observations on Benchmark Datasets}
The goal of this section is to observe the semi-supervised peaking phenomenon on several benchmark datasets (taken from \cite{Chapelle2006} and \cite{Lichman2013}) and relate these observations to the results in the previous sections. We generate semi-supervised learning curves for eight benchmark datasets as follows. We select $L=\lceil p/2 \rceil$ where $p$ is the dimensionality of the dataset after applying principal component analysis and retaining as many dimensions as required to retain $99\%$ of the variance. 

We then randomly, with replacement, draw additional training samples, with a maximum of $100$ for the smaller datasets and $1000$ for the larger datasets. We also sample a separate set of $1000$ objects with replacement to form the test set. The additional training samples are used as labeled examples by the supervised learner and as unlabeled samples for semi-supervised learning. We repeat this process $100$ times and average the results. These averaged learning curves are shown in \cref{fig:benchmark}.

\begin{knitrout}
\definecolor{shadecolor}{rgb}{0.969, 0.969, 0.969}\color{fgcolor}\begin{figure}
\includegraphics[width=\maxwidth]{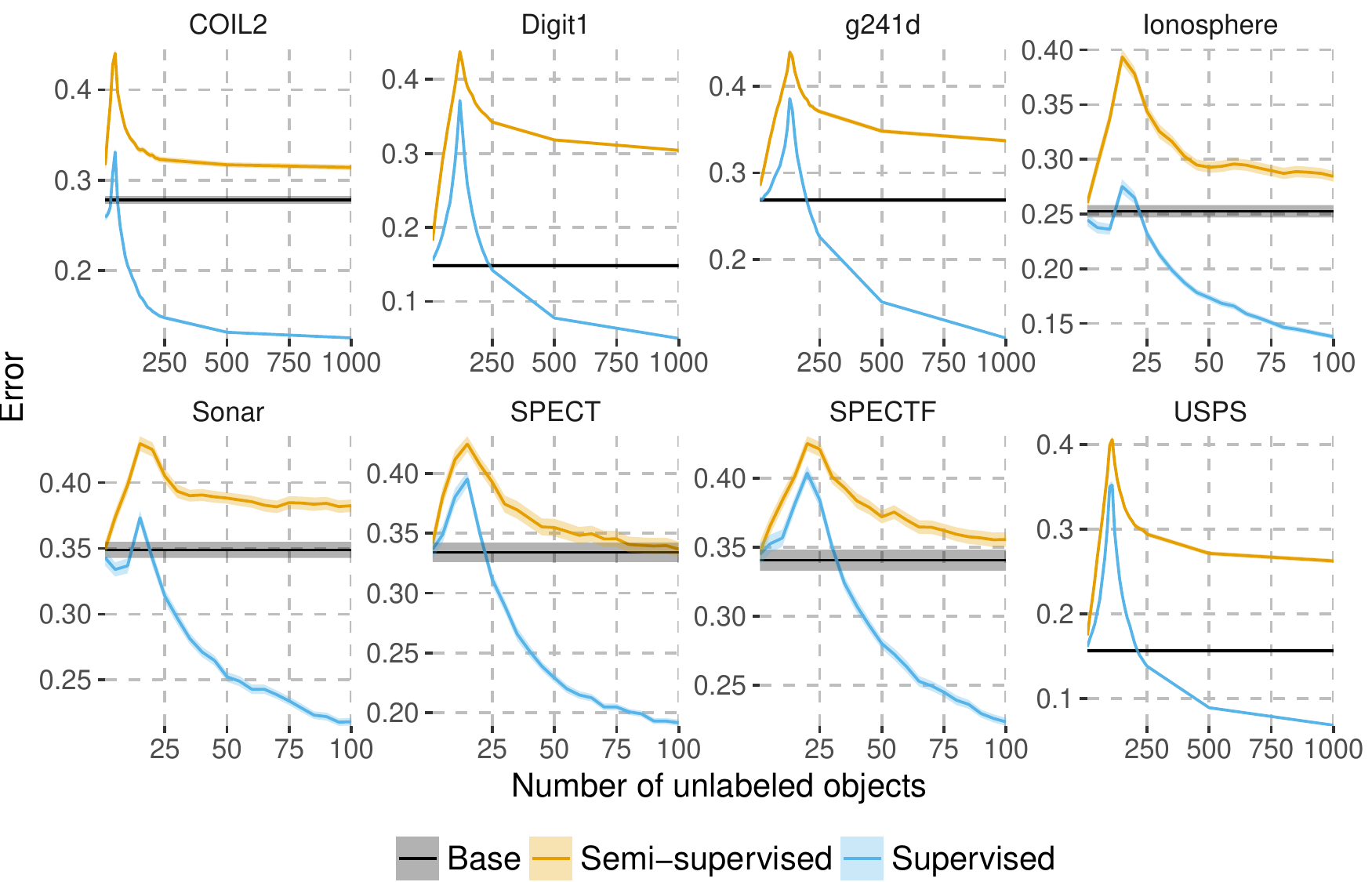} \caption[Learning curves on benchmark datasets]{Learning curves on benchmark datasets. The number of labeled objects is equal to $\lceil p/2 \rceil$. For the semi-supervised curve we add more unlabeled data, for the supervised curve more labeled data. ``Base'' shows performance of the supervised classifier using only the original $\lceil p/2 \rceil$ objects. For each dataset $100$ curves where generated and averaged. Shaded area (small) indicates the standard error around the mean.}\label{fig:benchmark}
\end{figure}

\end{knitrout}

Both behaviours studied in the previous sections, the steeper ascent in the semi-supervised setting before the peak and the slower decline after the peak, are apparent on these example datasets. We also notice that for most of these datasets it seems unlikely that the semi-supervised learning will improve over the base classifier. This may suggest we are in a scenario similar to the large difference between the means in \cref{fig:infinitedata}. The exception is the SPECT and SPECTF datasets, where the situation is more similar to the smaller $\delta$. Notice that for all datasets it is still possible we are in a situation similar to $\delta=2$ in \cref{fig:infinitedata}: while adding unlabeled data does not help with the given amount of labeled examples, this effect might reverse if a few more labeled objects become available.

\section{Discussion and Conclusion}
In this work, we have studied the behaviour of the learning curve for one particular semi-supervised adaptation of the least squares classifier. This adaptation, based on the ideas from \cite{Shaffer1991} and \cite{Fan2008}, was amenable to analysis. It is an open question what the typical learning curve for other semi-supervised least squares adaptations looks like, such as self-learning or the constraint based approach in \cite{Krijthe2015} where we first noticed this behaviour and which inspired us to look into this phenomenon. The lack of a closed form solution in these cases makes it more difficult to subject them to a similar analysis.  Nevertheless, the current study does provide insight in the additional problems that small samples entail in the semi-supervised setting and largely explains the learning curve behaviour, at least for the specific semi-supervised learner considered.

\subsubsection{Acknowledgements.}
This work was funded by project P23 of the Dutch public/private research network COMMIT.

\bibliographystyle{splncs03}
\bibliography{library}
\end{document}